**Title:** Chemistry42: An AI-based platform for *de novo* molecular design

**Authors:** Yan A. Ivanenkov[1], Alex Zhebrak[1], Dmitry Bezrukov[1], Bogdan Zagribelnyy[1], Vladimir Aladinskiy[1], Daniil Polykovskiy[1], Evgeny Putin[1], Petrina Kamya[1], Alexander Aliper[1], Alex Zhavoronkov[1]

**Affiliation:**

1. Insilico Medicine Hong Kong Ltd. Science Park East Avenue, Hong Kong Science Park, Pak Shek Kok, Hong Kong

**Corresponding author:** Alex Zhavoronkov, alex@insilico.com

**Abstract:** Chemistry42 is a software platform for *de novo* small molecule design that integrates Artificial Intelligence (AI) techniques with computational and medicinal chemistry methods. Chemistry42 is unique in its ability to generate novel molecular structures with predefined properties validated through *in vitro* and *in vivo* studies. Chemistry42 is a core component of Insilico Medicine's Pharma.ai drug discovery suite that also includes target discovery and multi-omics data analysis (PandaOmics) and clinical trial outcomes predictions (InClinico).

**Introduction**

Deep Learning (DL) has proven to be very effective in speech and image recognition. This is because DL-based architectures are uniquely suited for the automatic identification of patterns within complex, nonlinear data sets without the need for manual feature engineering. DL methods have successfully overcome limitations inherent in the standard techniques used for small molecule design (Chen et al. 2018; Vanhaelen, Lin, and Zhavoronkov 2020; Yang et al. 2019) which offers exciting possibilities for the development of new methods that efficiently explore uncharted chemical space.

Insilico Medicine, a company that develops AI algorithms for target discovery and generative chemistry, was one of the first groups to publish a method that uses a deep adversarial model for new compound generation (Kadurin, Aliper, et al. 2017). Since then, DL-based architectures that combine generative algorithms with reinforcement learning (RL) have been developed and applied in chemistry and pharmacology to generate novel molecular structures with predefined properties. Especially encouraging is the recent progress in the *de novo* design of active molecules that have been validated in both *in vitro* and *in vivo* assays (Zhavoronkov et al. 2019). The field of generative chemistry is now one of the fastest-growing drug discovery areas (Vanhaelen, Lin, and Zhavoronkov 2020; Schneider 2018; Merk et al. 2018). The Chemistry42™ platform has been routinely and successfully used at Insilico Medicine to drive the drug discovery process in several therapeutic areas. In the following sections, we describe the key features of the Chemistry42™ platform.

**Overview of the generative capabilities of the Chemistry42™ platform**

Chemistry42™ is a platform that connects state-of-the-art generative AI algorithms with medicinal chemistry expertise and best engineering practices. The main objective of this platform is to accelerate the design of novel molecules with user-defined properties. The general workflow for Chemistry42™ is illustrated and described in Figure 1.

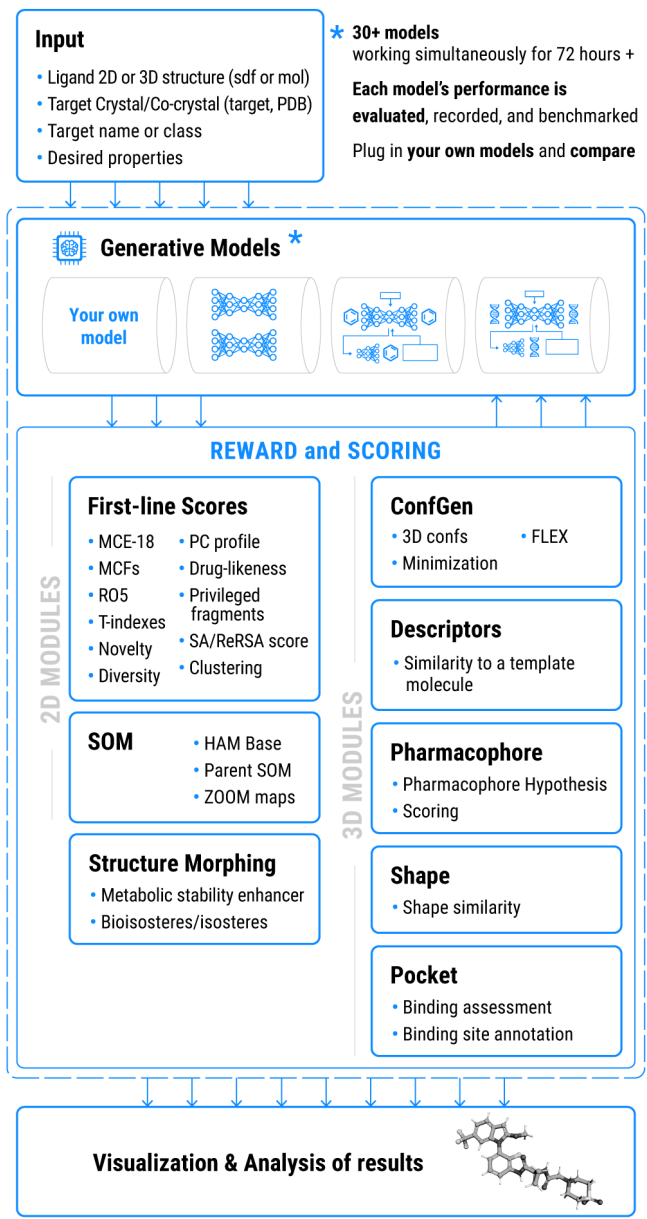

Figure 1. Schematic representation of the three-step workflow for a *de novo* generative experiment using the Chemistry42™ platform. In the first step, the user provides input and configures the generation experiment. Users can upload their data in multiple file formats and provide details on the desired properties of the generated structures. The second step involves deploying upto 30 generative models functioning in parallel to generate the novel structures, where users can specify which models should be enabled at this step. Power users can also choose to install additional custom generative model(s) and run them within the platform. All generated items undergo virtual screening based on a variety of filters for the generation phase. During the virtual

screening/optimization stage, multiple sets of reward and scoring functions, classified as either 2D or 3D modules, are used to dynamically assess the generated structures' properties according to the predefined criteria. These modules form the backbone of Chemistry42™'s RL-based generation procedure. Data from the virtual screening feeds back into the generative models to benchmark, optimize and provide valuable data to the models, which completes the cycle of reinforcement learning. The final step is analysis. The generated structures are automatically ranked according to customizable metrics based on their predicted properties, including synthetic accessibility, novelty, diversity, etc. The platform also provides users with interactive tools to monitor model performance.

Individual generative experiments are created as projects using the user-friendly web-based interface to Chemistry42™. They can be started using either ligand or structure-based drug design workflows (LBDD or SBDD approaches, respectively). The LBDD approach requires a 2D or 3D ligand structure as input in .sdf or .smi format. A pharmacophore hypothesis can also be added as needed. It can be created either manually using an external widget or automatically created within the platform. In the SBDD approach, the structure of a protein target, either in the *apo* format or in complex with a ligand, must be uploaded to the platform as a prepared .pdb file. One can pick either the pocket around the ligand (ligand binding site) or select one from the set of alternative pockets indicated by the *Pocket Scanner Module*. As with the case for LBDD, a pharmacophore hypothesis can also be added as needed (Figure 2). To complete the configuration of a generation experiment, the user defines acceptable ranges for multiple properties (e.g. physicochemical properties and diversity) of the generated structures. In both LBDD and SBDD approaches, advanced options enable the user to specify and fine-tune reward functions and which models should be used in an experiment. Once the configuration of the experiment is completed, the platform begins an interactive generative exploration of the chemical space using multiple generative models in parallel.

The generative procedure in Chemistry42 engages an asynchronous ensemble of proprietary generative models of different nature. These are carefully curated and selected algorithms with diverse architectures implementing distinct strategies. The platform utilizes multiple machine learning models and molecular representations for different scenarios to maximize their contribution and the platform's efficiency. For example, some of the models focus on the exploration of the chemical space, while others are tailored to improve these explored structures. In the current version of Chemistry42, there are over 30 generative models, including generative autoencoders (Polykovskiy et al. 2018; Zhavoronkov et al. 2019), generative adversarial networks (Kadurin, Nikolenko, et al. 2017; Kadurin, Aliper, et al. 2017; Putin et al. 2018), evolutionary algorithms, language models and others. Moreover, these models employ different molecular representations — string-based, graph-based, and 3D-based.

It is essential to understand and stimulate the interplay of multiple models. Rather than treating these algorithms as black-box solutions, we provide deep domain-specific analytics to understand the advantages and drawbacks of each approach. Combining various state-of-the-art machine learning methods, Chemistry42 is the most comprehensive generative chemistry effort to date, capable of delivering diverse, high-quality molecular structures in a fully automated fashion with speed. As the structures are generated, they are dynamically assessed using the reward and scoring functions in the platform.

Figure 2. Chemistry42™ interface for configuring an SBDD generative experiment

The reward and scoring functions used in Chemistry42 for the RL-based generation are classified as either two dimensional (2D) or three dimensional (3D) modules (Figure 1). The 2D modules are composed of multiple scores and in-house *Medicinal*

*Chemistry Filters* (*MCFs*) that are used to assess the generated structures. The *MCFs* include a set of over 460 in-house structure-based rules that exclude "bad" structures that contain structure alerts, PAINS (Baell and Holloway 2010) or functional groups that are reactive, unstable or potentially toxic. The *Medicinal Chemistry Evolution* (MCE-18) function is a unique molecular descriptor that scores structures by novelty in terms of their cumulative $sp^3$ complexity (Ivanenkov, Zagribelnyy, and Aladinskiy 2019). Other 2D modules include Lipinski's Rule of Five (RO5) (Lipinski et al. 2001), *Drug-likeness* and *T-indexes*, a rule-based filter that constitutes a set of rules to eliminate structures with an unbalanced number of carbons and heteroatoms. Similarity scores assess the 2D-similarity (cosine) between the generated structures and the reference dataset corresponding to the predefined and vast chemical space. The physicochemical profile of a compound is assessed using a set of molecular descriptors and predictive models in the *PhysChem Profile Predictor* module. *Drug-likeness* is estimated using a set of extended rules. The synthetic accessibility (SA) of the generated structures is assessed using the Retrosynthesis Related Synthetic Accessibility (ReRSA) score. ReRSA is an improved fragment-based SA estimation method that is based on the fragmentation of the generated structure from an organic synthesis perspective, which results in a more accurate estimation of SA. *Diversity* assessments and clustering metrics are performed for the generated structures using a combination of Finger-Print (FP)-based methods. Tracking the *Diversity* of the generated structures provides a means of understanding how structurally diverse they are based on the number of generated chemotypes following clustering. *Novelty* is estimated as (1−similarity), where similarity is calculated to the reported compounds from public sources (e.g. SureChembl). *Privileged Fragments* (PFs) are automatically defined structural motifs that contribute to the activity against a target or target class (Yet 2018). For each generated structure, this filter assesses the presence of PFs in the generated structures and returns a score. PFs can be specified by the user or generated automatically using the *Hierarchical Active Molecules (HAM) dataset* which contains biologically active molecules with reported *in vitro* activities against various targets organized hierarchically. The *HAM dataset* is also integrated into the platform's self-organizing maps (*SOM) Classifier Module* (Kohonen 2001) and ZOOM maps. *SOM Classifier Module* (general SOM map 100×100) is used to drive the generation towards the chemical space corresponding to a specified target class. Since the general SOM contains neurons with the classification power below the predefined threshold for a selected category of molecules, all the reference molecules from such neurons are collected and then subjected to automatically generated ZOOM maps of an adapted size to achieve a reliable classification accuracy. Structure Morphing module contains two components: a rule-based *Metabolic Stability Enhancer (Kirchmair et al. 2015)* to address metabolic instability and *Bioisostere Module* for bioisosteric/isosteric transformation (Brown 2012) to expand the generated chemical classes.

After assessing the generated structures with the 2D modules, the platform utilizes five different 3D modules for further assessment. The first 3D module generates a conformational ensemble for each generated structure (*ConfGen Module*). Conformational ensembles are generated through a combination of a set of rules and pre-defined substructure geometries based on small molecule or co-crystal X-ray data followed by energy minimization. A flexibility assessment (*FLEX score*) component is used to rank molecular structures by intrinsic rigidity. The second module evaluates the 3D similarity between the generated structures and a reference molecule (input ligand) using a set of calculated 3D-descriptors (*3D-Descriptors Module*). The third module is used to generate or construct a pharmacophore hypothesis(es) including all important binding points, distances, angles, and tolerance and automatically score the generated structures against the selected hypotheses (*Pharmacophore Module*). The fourth module (*Shape Similarity Module*) evaluates the 3D-shape similarity to a reference molecule using weighted Gaussian functions (Yan et al. 2013). The fifth module focuses on positioning and scoring the generated structures to assess how well they fit the selected binding site (*Pocket Module*).

All corresponding data including scores, molecular structures and model performance are stored and accessible on the results page of the platform where the generative experiment can be monitored in real-time till completion.

The average duration of a standard experiment that uses all generative models is 72 hours. For each generative model, the performance and convergence rate are monitored. This allows the user to follow the progress of their experiments in real time. The generated structures are automatically evaluated and ranked according to metrics incorporated in the modules that are integrated into the platform. Once a generative experiment is complete, the results can be analyzed through an interactive interface. The results can also be exported via an application programming interface (API).

**Model benchmarking**

With Chemistry42™, the user can compare the performance of all the generative models used in an experiment. The platform utilises a benchmarking system based on Molecular Sets (MOSES) system for assessing performance, reward components, novelty, diversity, and other metrics during the generation and after the experiment is completed (Polykovskiy et al. 2020). Based on the provided analytics, users can seamlessly compare the models' performance for every experiment. A record of the results and training data is kept during the experiment and stored to ensure that reproducibility and monitoring are both simple and feasible.

**Chemistry42™ interoperability**

Chemistry42™ is accessible through a user-friendly interface built on top of a distributed cloud platform with a scalable cloud architecture. The implementation integrates a variety of features aimed at optimizing its performance. This includes cluster management with Kubernetes, multiple flexible workflows, integrated monitoring and logging. The structure and interoperability of the Chemistry42™ platform allows its deployment on the cloud or on a user's own premises. For either deployment scenarios, the platform can be integrated into an already established workflow.

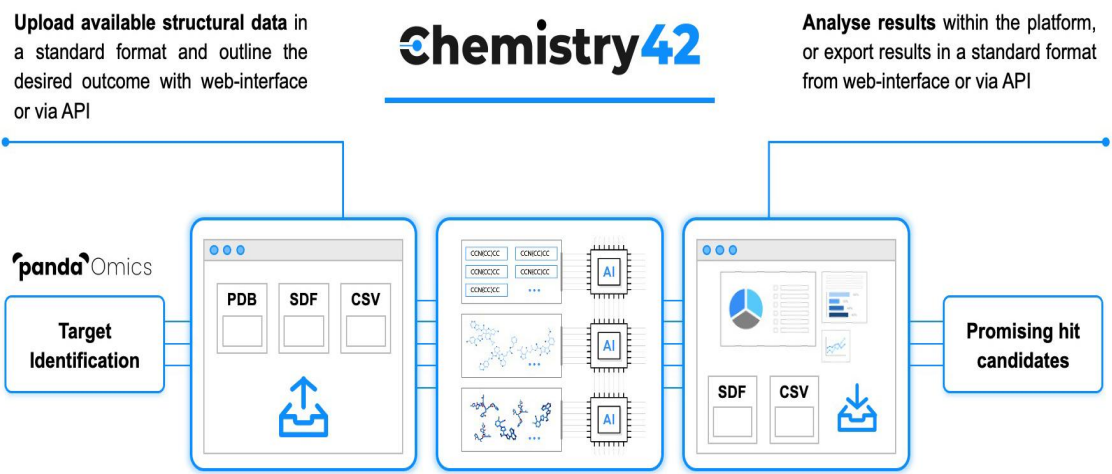

Figure 3. PandaOmics and Chemistry42™ platforms integrated into your drug discovery pipeline. The interoperability of these platforms allows an efficient interaction between target identification and de novo small molecule generation.

Chemistry42™ can be connected to Insilico Medicine's bioinformatics web service PandaOmics (https://pandaomics.com) (Figure 3). PandaOmics is a comprehensive computational suite for the analysis of -omics data that provides access to information ranging from disease signatures to prospective targets and existing drugs. PandaOmics combines classic bioinformatics methods with signaling pathway analysis using the iPANDA algorithm (Ozerov et al. 2016; Stamatas et al. 2017; Ravi et al. 2018; Saloura et al. 2019; Subbannayya et al. 2019). PandaOmics also provides access to an AI-powered toolkit including deep feature selection for pathway reconstruction, a pathway scoring engine, causal inference, deep-learned transcriptional response scoring engine and an activation-based scoring engine. This multimodal approach combines big data, chemistry, biology, and medicine and allows a complete characterization of the interplay between molecular structures, properties, alteration in biological samples and drug response required for target discovery.

**Conclusion**

The Chemistry42™ platform (https://insilico.com/chemistry42) is a customizable working environment that offers state-of-the-art AI technologies specifically developed for *de novo* molecular design. The flexible user-friendly interface makes

Chemistry42™ accessible to AI specialists, medicinal chemists, computational chemists, and other scientists working in the field of drug discovery. This unique collaborative feature will enable and foster relationships between different scientific communities and facilitate the decision-making process – a process which is exceptionally demanding in the field of drug design.

## Acknowledgements

The authors gratefully acknowledge the valuable comments and suggestions made by Dr. Jiye Shi from UCB Pharma (Slough, UK).

## Conflicts of Interest Disclosure

Y.A.I, A.Z., D.B., B.Z., V.A., D.P., E.P., P.K., A.A., A. Zhavoronkov work for Insilico Medicine, a commercial artificial intelligence company that developed the Chemistry42™ platform.